\documentclass{article}
\usepackage{journal}
\jmlrheading{}{2026}{}{}{}{A.~Altabaa, J.~Lafferty \& O.~Montasser}
\ShortHeadings{Optimal Autoregressive Chain-of-Thought Learning via Bagging}{Altabaa Lafferty Montasser}
\usepackage{amsmath,amssymb,amsthm,mathtools}
\usepackage{times}
\usepackage[utf8]{inputenc}
\usepackage[T1]{fontenc}
\usepackage{dsfont}
\usepackage{marvosym}
\usepackage{thmtools}
\usepackage{algorithm}
\usepackage[algo2e,ruled,linesnumbered,vlined]{algorithm2e}
\usepackage[noend]{algpseudocode}
\usepackage{microtype}
\usepackage[dvipsnames]{xcolor}
\usepackage{hyperref}
\usepackage{prettyref}
\usepackage[capitalise,nameinlink]{cleveref}
\usepackage{nicefrac}
\usepackage{enumitem}
\usepackage{booktabs,tabularx}
\hypersetup{colorlinks=true,citecolor=Maroon,linkcolor=Maroon,urlcolor=Maroon}
\setcitestyle{numbers,square}
\usepackage{comment}
\usepackage{mathrsfs}
\usepackage{tikz}
\usetikzlibrary{arrows.meta}

\newrefformat{alg}{\Cref{#1}}
\newrefformat{thm}{\Cref{#1}}
\newrefformat{corr}{\Cref{#1}}
\newrefformat{lem}{\Cref{#1}}
\newrefformat{prop}{\Cref{#1}}
\newrefformat{tab}{\Cref{#1}}
\newrefformat{app}{Appendix~\ref{#1}}

\newtheorem{lemma}{Lemma}

\newtheorem{proposition}{Proposition}
\theoremstyle{definition}
\newtheorem{remark}{Remark}

\newcommand{\X}{\mathcal X}
\newcommand{\Y}{\mathcal Y}

\newcommand{\Fc}{\mathcal F}

\newcommand{\E}{\mathds E}
\newcommand{\B}{\mathrm B}

\newcommand{\1}{\mathds 1}
\newcommand{\vc}{\mathrm{VCdim}}
\newcommand{\dVC}{d_{\mathrm{VC}}}
\newcommand{\dG}{d_{\mathrm{G}}}
\newcommand{\dN}{d_{\mathrm{N}}}
\newcommand{\dL}{d_{\mathrm{L}}}
\newcommand{\dDS}{d_{\mathrm{DS}}}
\newcommand{\eps}{\varepsilon}
\newcommand{\prn}[1]{\left(#1\right)}
\newcommand{\brk}[1]{\left[#1\right]}
\newcommand{\set}[1]{\left\{#1\right\}}
\newcommand{\ceil}[1]{\left\lceil#1\right\rceil}
\newcommand{\Ex}[1]{\underset{#1}{\E}}

\newcommand{\CONS}{\hyperref[eqn:erm]{\ensuremath{\textnormal{\textsc{Cons}}_{\mathrm{CoT}}}}}
\newcommand{\MAJ}{\mathrm{MAJ}}

\newcommand{\Unif}{\operatorname{Unif}}

\DeclareMathOperator*{\argmin}{arg\,min}
\newcommand{\ERMCoT}{\hyperref[eqn:erm-cot]{\ensuremath{\textnormal{\textsc{Erm}}_{\mathrm{CoT}}}}}
\newcommand{\Growth}[1]{\Gamma_{#1}}
\newcommand{\nU}{n_{\mathrm{u}}}

\definecolor{addcolor}{HTML}{1F5FAD}

\title{Optimal Autoregressive Chain-of-Thought Learning via Bagging}
\author{\hfill
       \begin{tabular}{c@{\hspace{2.5em}}c@{\hspace{2.5em}}c}
       \name Awni Altabaa & \name John Lafferty & \name Omar Montasser\\[4pt]
       \multicolumn{3}{c}{\normalsize\it Department of Statistics and Data Science, Yale University\textsuperscript{\normalfont\Letter}}
       \end{tabular}\hfill\null}

\begin{document}

\maketitle
\begin{NoHyper}{\renewcommand{\thefootnote}{\Letter}\footnotetext{\texttt{\{awni.altabaa, john.lafferty, omar.montasser\}@yale.edu}}}\end{NoHyper}

\begin{abstract}
    We study how to learn an autoregressive next-token generator $f\colon\Sigma^\star\to\Sigma$ from a class $\Fc$, given chain-of-thought (CoT) demonstrations of length $T$ generated by $f$. We present a simple algorithm based on bagging \citep{DBLP:journals/ml/Breiman96b}, which outputs a majority vote of generators that are trained to be CoT-consistent on bootstrap samples of the training data. For binary token sets, it achieves the optimal sample complexity of $O((\dVC+\log(1/\delta))/\eps)$, independent of the CoT length $T$, where $\dVC$ is the VC dimension of $\Fc$. For finite token sets $\Sigma$, the VC dimension is replaced by a multiclass analogue, the Graph dimension $\dG$, and the sample complexity is optimal up to a $\log|\Sigma|$ factor. The number of generators in the majority-vote ensemble is $O(\log(T/\delta))$, and we show that the ensemble can be distilled into a \emph{single} generator in $\Fc$ at the expense of $n\log(nT)$ \emph{unlabeled} prompts.
\end{abstract}

\section{Introduction}
\label{sec:intro}

Large language models have become one of the most important technologies of recent years, yet the mechanism at their core is strikingly simple: a single next-token predictor is applied repeatedly, and each token the model produces is appended to the context of the next prediction. Such models are most successful when they do not answer immediately, but first generate a sequence of intermediate steps, a \emph{chain-of-thought}, from which the answer follows \citep{NyeAGMABDLBLSO21,WeiWSBIXCLZ22}. Training on demonstrations of such step-by-step reasoning has become a central ingredient of modern training pipelines. This success raises questions that empirical progress alone cannot settle. How many demonstrations are needed to learn to reason correctly? Does a longer reasoning process make learning harder? And which learning algorithms make the best use of the available data? Addressing these questions calls for a theory of learning tailored to autoregressive generation.
 
\paragraph{An Autoregressive Learning Model.}
In the spirit of classical PAC learning theory \citep{VapnikChervonenkis71,DBLP:journals/cacm/Valiant84,DBLP:journals/jacm/BlumerEHW89,vapnik1998statistical},
Joshi et al.~\citep{pmlr-v291-joshi25a} introduced an elegant PAC-style framework for studying autoregressive learning, which we adopt throughout. Here, a language model is abstracted as an autoregressive next-token generator $f\colon\Sigma^\star\to\Sigma$ from a class $\Fc$, where $\Sigma$ is a set of tokens. On a prompt $x\in\Sigma^\star$, the generator produces the token $f^{(1)}(x)=f(x)$, then $f^{(2)}(x)=f(x\circ f^{(1)}(x))$, and so on; after $T$ steps it has produced the chain-of-thought $f^{(1:T)}(x)=(f^{(1)}(x)\circ \cdots \circ f^{(T)}(x))\in\Sigma^T$, whose last token is interpreted as the answer. The same generator is used at every step, exactly as a language model applies the same weights to every prefix. Two forms of supervision arise naturally. Under \emph{end-to-end supervision}, the learner observes prompts with their final answers, while the reasoning that produced them remains hidden. Under \emph{chain-of-thought supervision}, the subject of this paper, the learner observes the entire chain-of-thought.

\enlargethispage{\baselineskip}Let $P$ be a distribution over $\Sigma^\star\times\Sigma^T$ that is \emph{realizable} by some $f_\star\in\Fc$, in the sense that $\Pr[f_\star^{(1:T)}(x)=y]=1$. The learner observes $n$ i.i.d.~samples $S=((x_j,y_j))_{j=1}^n\sim P^n$ and outputs a predictor $\widehat f\colon\Sigma^\star\to\Sigma$, which is then run autoregressively. Its performance is measured by the CoT risk, where the predictor $\widehat{f}$ must reproduce the entire chain-of-thought, and a single incorrect token renders the output incorrect:\footnote{Under chain-of-thought supervision, Joshi et al.~\citep{pmlr-v291-joshi25a} and Hanneke, Mehalel, and Moran~\citep{HannekeMehalelMoran2026} establish guarantees on the end-to-end risk $\Pr_{(x,y)\sim P}[\widehat f^{(T)}(x)\neq y_T]$, which is at most the CoT risk~\eqref{eqn:cot-risk}. Their analyses proceed by bounding the CoT risk. The focus of this paper is the CoT risk, which is distinct from the end-to-end risk, and learning guarantees for the two can differ; see \citep{NEURIPS2025_235a9abb} and \citep[][Section~8]{HannekeMehalelMoran2026} for further discussion.}
\begin{equation}
    \label{eqn:cot-risk}
    \Pr_{(x,y)\sim P}\brk{\exists t\in[T]:\ \widehat f^{(t)}(x)\neq y_t}.
\end{equation}
The goal is to determine the sample complexity of this problem as a function of the accuracy $\eps$, the confidence $\delta$, the CoT length $T$, and the complexity of the class $\Fc$. Moreover, this sample complexity should be attained with simple and efficient learning algorithms. When $T=1$, the problem is precisely multiclass learning of $\Fc$ with label set $\Sigma$. Learning with chain-of-thought supervision is therefore at least as hard as multiclass learning; a central question is whether learning to produce a long chain-of-thought is harder than learning to predict a single token.
 
\begin{table}[t]
    \centering
    \small
    \renewcommand{\arraystretch}{1.25}
    \setlength{\tabcolsep}{4pt}
    \begin{tabularx}{\linewidth}{@{}l c l >{\raggedright\arraybackslash}X@{}}
        \toprule
        \textbf{Reference} & \textbf{Sample Complexity} & \textbf{Alphabet} & \textbf{Learner} \\
        \midrule
        \citep{pmlr-v291-joshi25a} & $O\bigl((d\log(2T)\log(2/\eps)+\log(1/\delta))/\eps\bigr)$ & binary & CoT-consistent learner \\
        \addlinespace[5pt]
        \citep{HannekeMehalelMoran2026} & $O\bigl((dd^\star\log(2dd^\star)\log(2/\eps)+\log(1/\delta))/\eps\bigr)$ & binary & sample-compression majority-vote \\
        \addlinespace[5pt]
        \citep{DoronAradMehalelMossel2026} & $O\bigl((\dL+\log(1/\delta))/\eps\bigr)$ & arbitrary & online-to-batch conversion \\
        \addlinespace[5pt]
        \citep{Li2026Autoregressive} & $O\bigl((\dDS+\log(1/\delta))/\eps\bigr)$ & arbitrary & one-inclusion hypergraphs \\
        \addlinespace[5pt]
        \textbf{This work} (\prettyref{thm:linearvc}) & $O\bigl((\dG+\log(1/\delta))/\eps\bigr)$ & arbitrary & Bagging CoT-consistent predictors \\
        \midrule
        \mbox{Lower bound \citep{Li2026Autoregressive}} & $\Omega\bigl((\dDS+\log(1/\delta))/\eps\bigr)$ & arbitrary & any learner, already for $T=1$ \\
        \bottomrule
    \end{tabularx}
    \caption{Sample complexity of realizable learning with chain-of-thought supervision. Here $d$, $\dG$, $\dL$, and $\dDS$ are the VC, Graph, Littlestone, and DS dimensions of $\Fc$, and $d^\star$ is its dual VC dimension. We refer the reader to \prettyref{sec:prelim} for definitions and background on these complexity measures.}
    \label{tab:comparison}
\end{table}

\paragraph{Prior Work.}
For context, we summarize existing sample complexity bounds for autoregressive learning under chain-of-thought supervision. These bounds are summarized in \prettyref{tab:comparison}, and complexity measures are reviewed in \prettyref{sec:prelim}; a broader discussion of related work is deferred to \prettyref{sec:related}. The story begins with Joshi et al.~\citep{pmlr-v291-joshi25a}, who showed that a single call to a $\CONS$ oracle, which returns a generator that is CoT-consistent with the training data, achieves a sample complexity of order $d\log T$, where $d$ is the VC dimension of $\Fc$. Then, Hanneke, Mehalel, and Moran~\citep{HannekeMehalelMoran2026} showed that this $\log T$ factor can be removed, with a majority-vote compression scheme \citep{DBLP:journals/jacm/MoranY16}, but at the price of a factor of the dual VC dimension $d^\star$, which may be exponentially larger than $d$ \citep{Assouad83,KleerS23}. Afterwards, Doron-Arad, Mehalel, and Mossel~\citep{DoronAradMehalelMossel2026} obtained another $T$-independent bound through online learning, but in terms of the Littlestone dimension $\dL$, which can be infinite even for simple classes (e.g., thresholds on the line).

Most recently, Li~\citep{Li2026Autoregressive} settled the optimal sample complexity of the problem: every autoregressive class $\Fc$ is learnable with sample complexity $O((\dDS+\log(1/\delta))/\eps)$, independently of $T$, where $\dDS$ is the Daniely--Shalev-Shwartz (DS) dimension of $\Fc$ \citep{DBLP:conf/colt/DanielyS14}, which characterizes multiclass learnability \citep{DBLP:conf/focs/BrukhimCDMY22} (see \prettyref{sec:prelim} for definitions). This result builds on recent progress in multiclass learning, which established that the DS dimension also governs the optimal sample complexity of multiclass learning \citep{DBLP:journals/corr/abs-2604-12952,DBLP:journals/corr/abs-2604-24749}. As the case $T=1$ is itself multiclass learning, this bound cannot be improved in general. Thus, information-theoretically, generating a long chain-of-thought is no harder than generating a single token. The learner achieving this sample complexity, however, is far from practical. It is based on the one-inclusion graph \citep{DBLP:journals/iandc/HausslerLW94}. Given a test prompt $x$, the learner builds the one-inclusion graph, whose vertices are all the chains-of-thought that functions in $\Fc$ can generate on the training prompts and on $x$, orients its edges, and reads off its prediction from the orientation. This graph is typically enormous, and it must be rebuilt and reoriented for every new test prompt, making the learner computationally intractable, and in sharp contrast to how language models are trained and deployed in practice. This raises a natural question that we study in this paper:
\begin{center}
    \emph{Is there a simple and efficient algorithm for autoregressive learning\\ that attains the optimal sample complexity?}
\end{center}

\section{Our Contributions}
\label{sec:contributions}
We answer this question affirmatively for binary alphabets, and up to a factor of $\log|\Sigma|$ for finite alphabets, with one of the oldest and simplest ideas in machine learning:~bagging. Introduced by Breiman~\citep{DBLP:journals/ml/Breiman96b}, bagging trains a base learner on bootstrap samples, drawn with replacement from the training set, and combines the resulting predictors by a majority vote. Our algorithm, bagging CoT-consistent predictors (\prettyref{alg:bagging}), carries this idea over to autoregressive learning: it trains $N=O(\log(T/\delta))$ CoT-consistent predictors on bootstrap samples, and generates each token of the chain-of-thought by a majority vote of their next-token predictions. We show that it learns every autoregressive class $\Fc$ with sample complexity $O((\dG+\log(1/\delta))/\eps)$, independent of the CoT length $T$, where $\dG$ is the Graph dimension of $\Fc$ (\prettyref{thm:linearvc}). For binary alphabets, the Graph and DS dimensions both coincide with the VC dimension, so our bound matches the optimal sample complexity of \citep{Li2026Autoregressive}. For finite alphabets, our bound is optimal up to a factor of $\log|\Sigma|$. The output of \prettyref{alg:bagging} is an ensemble of $N$ generators, which multiplies the inference cost per token by $N$. We further show that the ensemble can be distilled into a single generator in $\Fc$ using $O(n \log(nT))$ \emph{unlabeled} prompts while retaining the labeled sample complexity (see \prettyref{prop:distill}).

\refstepcounter{algocf}
\label{alg:bagging}
{\setlength{\fboxsep}{4pt}
    \begin{center}
        \fbox{\begin{minipage}{0.92\linewidth}
                {
                    \centering \textbf{Algorithm~\thealgocf: Bagging CoT-Consistent Predictors}\\
                }
                \textbf{Input:} Training set $S=((x_j, y_j))_{j=1}^n$, confidence $\delta$, and a $\CONS$ oracle \eqref{eqn:erm}.\par
                1. Set $N=\max\set{4,\ceil{\log_2(2T/\delta)}}$, and $J_n=\set{\ceil{n/4},\dots,n-1}$.\par
                2. For each $i=1,\dots,N$:\par
                \hspace*{1.25em}3. Sample $k$ uniformly from $J_n$.\par
                \hspace*{1.25em}4. Sample a bootstrap $S'_i$ of $k$ samples uniformly with replacement from $S_{\leq k}=((x_j, y_j))_{j=1}^{k}$.\par
                \hspace*{1.25em}5. Run the $\CONS$ oracle on $S'_{i}$, and let $\widehat{f}_{S'_i}=\CONS(S'_i)$ be the returned predictor.\par
                \textbf{Output:} The majority-vote predictor over $\widehat f_{S'_1},\dots,\widehat f_{S'_N}$, defined as follows (with fixed tie-breaking):
                {\vskip -2.5mm}
                \[
                    \begin{aligned}
                        \widehat y_1&=\MAJ(\widehat f_{S'_1},\dots,\widehat f_{S'_N})(x)
                        =\underset{y\in\Sigma}{\arg\max}~~{\textstyle\sum_{i=1}^N}\1\{\widehat f_{S'_i}(x)=y\},\\
                        \widehat y_t&=\MAJ(\widehat f_{S'_1},\dots,\widehat f_{S'_N})(x\circ\widehat y_{1:t-1}),\qquad 2\leq t\leq T.
                    \end{aligned}
                \]
        \end{minipage}}
        \smallskip
    \end{center}
}

\begin{restatable}{theorem}{realizablethm}
    \label{thm:linearvc}
    For any autoregressive class $\Fc$ with Graph dimension $\dG$, any $\CONS$ oracle, and any distribution $P$ over $\Sigma^\star\times\Sigma^T$ such that $P\brk{f_\star^{(1:T)}(x)=y}=1$ for some $f_\star\in\Fc$, for every $n\geq 4$ and every $\delta \in (0,1)$, letting $N=\max\set{4,\ceil{\log_2(2T/\delta)}}$, with probability at least $1-\delta$ over the random draw of a training dataset $S=\prn{\prn{x_j, y_j}}_{j=1}^{n}\sim P^n$ and over the $N$ bootstrap samples $S'_1,\dots,S'_N$ drawn as in \prettyref{alg:bagging},
    \[
        \Pr_{(x,y)\sim P}\brk{ \exists t\in[T]: \MAJ(\widehat{f}_{S'_1},\dots, \widehat{f}_{S'_N})^{(t)}(x) \neq y_t} = O\prn{\frac{\dG}{n} + \frac{1}{n}\log\prn{\frac{1}{\delta}}}.
    \]
\end{restatable}

\begin{remark}[Optimality]
    The sample complexity in \prettyref{thm:linearvc} compares with known lower bounds as follows.
    \begin{itemize}[leftmargin=1.5em, itemsep=1pt, topsep=1pt]
        \item \emph{Binary alphabets.} The bound is optimal since $\dG$ is exactly the VC dimension of $\Fc$, and this matches the classical lower bound $\Omega((\dG+\log(1/\delta))/\eps)$ for PAC learning \citep{DBLP:journals/iandc/EhrenfeuchtHKV89}, which holds already for $T=1$.
        \item \emph{Finite alphabets.} The bound is optimal up to a factor of $\log|\Sigma|$: the Graph dimension satisfies $\dG\leq 4.67\dN\log_2|\Sigma|$, where $\dN$ is the Natarajan dimension of $\Fc$ \citep{DBLP:journals/ml/Natarajan89, DBLP:journals/jcss/Ben-DavidCHL95,DBLP:journals/jmlr/DanielySBS15}, while multiclass learning, and hence our problem, requires $\Omega((\dN+\log(1/\delta))/\eps)$ samples \citep{DBLP:journals/ml/Natarajan89,DBLP:journals/jmlr/DanielySBS15}.
    \end{itemize}
\end{remark}

\begin{remark}
    The same guarantee holds with $N=O(d^\star+\log(1/\delta))$ bootstrap samples (and oracle calls), where $d^\star$ is the dual VC dimension of $\Fc$ (see \prettyref{lem:sparsification}). Hence $N=O\prn{\log(1/\delta)+\min\set{d^\star,\log T}}$ suffices.
\end{remark}
 
\prettyref{alg:bagging} is simple and efficient. It accesses the class only through a $\CONS$ oracle, which it calls $N=O(\log(T/\delta))$ times, and its output is itself a next-token generator: each token is obtained by evaluating the $N$ base predictors on the current prefix and returning the most frequent prediction, so a chain-of-thought of length $T$ costs only $NT$ evaluations of predictors in $\Fc$. We also remark that, like bagging in general \citep{DBLP:journals/ml/Breiman96b}, both training and inference in \prettyref{alg:bagging} can be \emph{parallelized} across the $N$ base predictors. Moreover, the ensemble can be distilled into a single generator in $\Fc$ after training, removing the factor of $N$ in the cost per token (see \prettyref{sec:distillation}). By contrast, the optimal learner of Li~\citep{Li2026Autoregressive} predicts by building and orienting a one-inclusion graph over all chains-of-thought that $\Fc$ can generate on the training prompts and the test prompt, a construction that must be repeated for every new test prompt and is computationally intractable.

For very large (and infinite) alphabets, a natural question is whether a simple algorithm such as \prettyref{alg:bagging}, which aggregates predictors in $\Fc$, can attain the optimal sample complexity $O((\dDS+\log(1/\delta))/\eps)$. The answer is unfortunately \emph{no}. The obstruction is inherited from multiclass learning, and therefore arises already for $T=1$, where our problem reduces to multiclass learning of $\Fc$. In our context, recent impossibility results for aggregations of proper learners \citep{AsilisHV25} have two consequences. First, there are classes $\Fc$ on which majorities of consistent learners require $\Omega((\dG+\log(1/\delta))/\eps)$ samples, so the dependence on the Graph dimension in \prettyref{thm:linearvc} is tight for \prettyref{alg:bagging}. Second, there is a class $\Fc$ of finite DS dimension that cannot be learned to constant error by any algorithm that aggregates finitely many predictors from the class, so for infinite alphabets, no such algorithm can attain the optimal sample complexity. This perhaps suggests a fundamental tension between the simplicity of a learning algorithm and the optimality of its sample complexity, in line with recent work on the limits of simple learning principles for multiclass learning \citep{AsilisDSSTW26}.

\section{Overview of Proof Ideas}
\label{sec:overview}

We highlight the main ideas behind the proof of \prettyref{thm:linearvc}, after briefly describing the prior approaches. Joshi et al.~\citep{pmlr-v291-joshi25a} analyze a CoT-consistent predictor through the VC dimension of the class of chain-of-thought errors, which incurs a factor of $\log T$. Hanneke, Mehalel, and Moran~\citep{HannekeMehalelMoran2026} proceed with an analysis based on sample compression schemes, which removes the dependence on $T$ but incurs a factor of the dual VC dimension $d^\star$. Li~\citep{Li2026Autoregressive} predicts by orienting one-inclusion graphs of the class of chains-of-thought, which attains the optimal sample complexity but is computationally intractable. We take a different route, inspired by the recent analysis of bagging for adversarially robust learning in \citep{Montasser26}: rather than analyzing a single predictor, we analyze the distribution over CoT-consistent predictors trained on random samples, and aggregate them by bagging. Throughout, let $\Fc$ be a class with Graph dimension $\dG$, fix an arbitrary $\CONS$ oracle, and let $P$ be a realizable distribution. The analysis has three main parts.

\paragraph{Distribution of CoT-Consistent Predictors.} The key idea is to analyze the distribution of the CoT-consistent predictors $\widehat f_S$ obtained from random training samples $S\sim P^n$. \prettyref{lem:key} shows that, for a random test example, the fraction of these predictors that err at the worst step along the target chain-of-thought is small:
\begin{equation}
    \label{eqn:overview-distribution}
    \Ex{(x,y)\sim P}\brk{\max_{t\in[T]}\ \Pr_{S\sim P^n}\brk{\widehat f_S(x\circ y_{1:t-1})\neq y_t}^2}=O\prn{\frac{\dG}{n}}.
\end{equation}
In words: on most test examples, at every step of the target chain-of-thought, the correct token is predicted by most CoT-consistent predictors, even though the predictors that err may change from one step to the next. This is a weaker requirement than asking a single predictor to be correct at all steps at once, and it is the reason the bound does not depend on $T$. For a single CoT-consistent predictor, Li~\citep[][Theorem~6.2]{Li2026Autoregressive} shows that a dependence on $T$ is unavoidable: for a suitable binary class and $\CONS$ oracle,
\[
    \Ex{(x,y)\sim P}\Ex{S\sim P^n}\brk{\max_{t\in[T]}\1\set{\widehat f_S(x\circ y_{1:t-1})\neq y_t}}=\Omega\prn{\frac{\dG\log T}{n}}.
\]
The proof of \eqref{eqn:overview-distribution} rests on a simple observation. For each test example, we fix in advance the step at which a random predictor $\widehat f_S$ is most likely to err; this choice depends on $P$, $n$, and the $\CONS$ oracle, but not on the sample. Every CoT-consistent predictor is correct at every step of every training chain-of-thought, and in particular at the chosen step of each training example. Bounding \eqref{eqn:overview-distribution} thus amounts to bounding the expected number of mistakes of a consistent learner on a fixed next-token prediction problem, over a class of error regions of VC dimension at most $\dG$, a standard analysis. To avoid logarithmic factors, we bound the second moment in \eqref{eqn:overview-distribution} directly, relying on recent second-moment bounds for consistent learners \citep{DBLP:conf/colt/Aden-AliHLZ24,DBLP:journals/corr/abs-2606-13614}; bounding only the first moment, as in \citep{DBLP:journals/jacm/BlumerEHW89}, would incur an extra multiplicative factor of $\log(n/\dG)$.

\paragraph{Bagging Leave-One-Out Analysis.} Bound \eqref{eqn:overview-distribution} concerns predictors trained on fresh samples from $P$, whereas \prettyref{alg:bagging} trains predictors on bootstraps of a single sample $S$. To connect the two, we consider an idealized bagging predictor, which aggregates the CoT-consistent predictors trained on all ordered bootstraps of $S$. For $x'\in\Sigma^\star$ and $y'\in\Sigma$, its bagging score is the fraction of these predictors that predict $y'$ on $x'$:
\[
    \widehat{\B}_S(y'\mid x'):=\Pr_{S'\sim\Unif(S)^{|S|}}\brk{\widehat f_{S'}(x')=y'}.
\]
We bound its leave-one-out error. The key observation is that a bootstrap of $S$ of size $n-1$ omits the $i$-th example with probability $(1-1/n)^{n-1}\geq 1/e$, and is then a bootstrap of the sample $S_{-i}$ without it. This observation underlies the out-of-bag estimate of the error of bagging \citep{Breiman96OOB,WolpertM99,Tibshirani96} and, more broadly, leave-one-out bootstrap estimates \citep{Efron83,EfronT97}; here, we use it to bound the leave-one-out error rather than to estimate it. Since bootstraps of $S$ are samples from its empirical distribution $D:=\Unif(S)$, the observation reduces the leave-one-out error to the quantity bounded in \eqref{eqn:overview-distribution}, for $D$ in place of $P$ (\prettyref{lem:leave-one-out-margin}):
{\small
\begin{equation}
    \label{eqn:overview-loo}
    \frac{1}{n}\sum_{i=1}^n\max_{1\leq t\leq T}\prn{1-\widehat{\B}_{S_{-i}}(y_{i,t}\mid x_i\circ y_{i,1:t-1})}^2\leq e^2\Ex{(x,y)\sim D}\brk{\max_{1\leq t\leq T}\Pr_{S'\sim D^{n-1}}\brk{\widehat f_{S'}(x\circ y_{1:t-1})\neq y_t}^2}.
\end{equation}}%
By \eqref{eqn:overview-distribution}, the right-hand side is $O(\dG/n)$. Finally, by exchangeability (\Cref{rem:expectationbound}), \eqref{eqn:overview-loo} also bounds by $O(\dG/n)$ the expected squared worst-step error of the idealized bagging predictor on a fresh example. Since its majority vote errs on a chain-of-thought only if the correct token receives at most half of the votes at some step, its expected CoT risk is $O(\dG/n)$ as well, independently of $T$.

\paragraph{High-probability Bound and Sparsification.} To obtain a high-probability bound with the optimal dependence on $\log(1/\delta)$, we employ the suffix-averaging technique of Aden-Ali et al.~\citep{DBLP:conf/focs/Aden-AliCSZ23}, which converts leave-one-out bounds for permutation-invariant predictors into high-probability bounds. This leads to the suffix-bagging predictor, which averages the bagging predictors of the prefixes $S_{\leq k}$ of the sample, and whose squared worst-step error is $O((\dG+\log(1/\delta))/n)$ with high probability (\prettyref{lem:full-agg}).

The main challenge is that the suffix-bagging predictor is intractable: it aggregates the predictors trained on all ordered bootstraps of every prefix, exponentially many in $n$, each requiring a call to the $\CONS$ oracle. A key idea in \prettyref{alg:bagging} is to randomly sample $N$ of these predictors and take their majority vote, which can be thought of as a Monte Carlo approximation of the suffix-bagging predictor. In the spirit of \citep[][Lemma~3]{DBLP:conf/colt/Larsen23}, we show that $N=O(\log(T/\delta))$ samples suffice, independently of $n$ (\prettyref{lem:sparsification-risk}). The reason is that only the votes on the $T$ target prefixes of each example matter: at each of them, the sampled majority errs with probability exponentially small in $N$ and proportional to the squared error controlled by \prettyref{lem:full-agg}, and a union bound over the $T$ steps concludes. This is the only place where the CoT length appears, and it is the reason we measure errors by the squared worst-step error throughout: a thresholded loss would only yield an additive approximation error, and thus a number of oracle calls growing logarithmically with $n$. Alternatively, the classical sparsification argument of Moran and Yehudayoff~\citep{DBLP:journals/jacm/MoranY16} gives $N=O(d^\star+\log(1/\delta))$, independently of $T$ (\prettyref{lem:sparsification}), but the dual VC dimension $d^\star$ can be exponentially larger than $\dG$ \citep{Assouad83,KleerS23}.

\section{Technical Definitions}
\label{sec:prelim}

\paragraph{Autoregressive Learning.} We recall the setting of \prettyref{sec:intro} and fix notation. Let $\Sigma$ be a set of tokens, finite or infinite, let $\Sigma^\star$ be the set of finite strings over $\Sigma$, and write $[r]=\set{1,\dots,r}$ for each positive integer $r$. For a generator $f\colon\Sigma^\star\to\Sigma$ and a prompt $x\in\Sigma^\star$, the generated tokens are defined recursively by
\[
    f^{(1)}(x)=f(x)
    \qquad\text{and}\qquad
    f^{(t)}(x)=f\prn{x\circ f^{(1)}(x)\circ\cdots\circ f^{(t-1)}(x)}\quad\text{for } t\geq 2,
\]
where $\circ$ is concatenation, and $f^{(1:T)}(x)=f^{(1)}(x)\circ\cdots\circ f^{(T)}(x)\in\Sigma^T$. The learner receives $S=((x_j,y_j))_{j=1}^n\sim P^n$ from a distribution $P$ over $\Sigma^\star\times\Sigma^T$ realizable by $\Fc$, and is evaluated by its CoT risk \eqref{eqn:cot-risk}.

\paragraph{CoT-Consistent Oracle.} Our algorithms access $\Fc$ only through a $\CONS$ oracle: a deterministic rule that, given a sample $S$ realizable by $\Fc$, returns $\widehat f_S:=\CONS(S)\in\Fc$ reproducing every chain-of-thought in $S$:
\begin{equation}
    \label{eqn:erm}
    \widehat{f}_S :=\CONS(S) \in\set{f\in\Fc:\ f^{(t)}(x_j)=y_{j,t}\ \text{for all } j\in[n] \text{ and } t\in[T]}.
\end{equation}
We call such a generator \emph{CoT-consistent} with $S$. Equivalently, it correctly predicts every next token along the chains-of-thought in $S$: $\widehat f_S(x_j\circ y_{j,1:t-1})=y_{j,t}$ for all $j\in[n]$ and $t\in[T]$.

\paragraph{Complexity Measures.} We briefly review four standard combinatorial dimensions of $\Fc$, viewed as a class of next-token predictors from $\Sigma^\star$ to $\Sigma$: the VC, Graph, Natarajan, and DS dimensions. For prompts $x_1,\dots,x_k\in\Sigma^\star$, write $\Fc|_{(x_1,\dots,x_k)}=\set{(f(x_1),\dots,f(x_k)):f\in\Fc}\subseteq\Sigma^k$ for the set of next-token patterns that $\Fc$ realizes on them. Each dimension below is the largest $k$ for which some prompts $x_1,\dots,x_k$ are shattered in the corresponding sense, or $\infty$ if no largest $k$ exists.
\begin{itemize}[leftmargin=1.5em, itemsep=2pt, topsep=3pt]
    \item \emph{VC dimension} $\dVC(\Fc)$ \citep{VapnikChervonenkis71}, for $|\Sigma|=2$: $x_1,\dots,x_k$ are shattered if $\Fc|_{(x_1,\dots,x_k)}=\Sigma^k$.
    \item \emph{Graph dimension} $\dG(\Fc)$ \citep{DBLP:journals/ml/Natarajan89}: $x_1,\dots,x_k$ are G-shattered if there are tokens $s_1,\dots,s_k\in\Sigma$ such that, for every $B\subseteq[k]$, some $f\in\Fc$ satisfies $f(x_i)=s_i$ exactly for $i\in B$. Equivalently, $\dG(\Fc)$ is the VC dimension of the 0-1 loss class $\set{(x,y)\mapsto\1\set{f(x)\neq y}:f\in\Fc}$ over $\Sigma^\star\times\Sigma$.
    \item \emph{Natarajan dimension} $\dN(\Fc)$ \citep{DBLP:journals/ml/Natarajan89}: $x_1,\dots,x_k$ are N-shattered if there are tokens $s_i\neq s'_i$, for $i\in[k]$, such that, for every $B\subseteq[k]$, some $f\in\Fc$ satisfies $f(x_i)=s_i$ for $i\in B$ and $f(x_i)=s'_i$ for $i\notin B$.
    \item \emph{DS dimension} $\dDS(\Fc)$ \citep{DBLP:conf/colt/DanielyS14, DBLP:conf/focs/BrukhimCDMY22}: $x_1,\dots,x_k$ are DS-shattered if $\Fc|_{(x_1,\dots,x_k)}$ contains a $k$-dimensional pseudo-cube, that is, a nonempty finite set $H\subseteq\Sigma^k$ in which every pattern $h\in H$ has, for each $i\in[k]$, a neighbor $h'\in H$ with $h'_i\neq h_i$ and $h'_j=h_j$ for all $j\neq i$.
\end{itemize}
When the class is clear from context, we simply write $\dVC$, $\dG$, $\dN$, and $\dDS$. For binary alphabets, all four notions coincide: $\dG=\dN=\dDS=\dVC$. For finite alphabets, they are related by
\[
    \dN\leq\dDS\leq\dG\leq 4.67\,\dN\log_2|\Sigma|.
    \qquad\text{\citep{DBLP:journals/jcss/Ben-DavidCHL95,DBLP:journals/jmlr/DanielySBS15,DBLP:conf/focs/BrukhimCDMY22}}
\]
In particular, the three dimensions are finite simultaneously, and both the Natarajan and the DS dimensions characterize learnability for finite alphabets \citep{DBLP:journals/jmlr/DanielySBS15,DBLP:conf/focs/BrukhimCDMY22}. For infinite alphabets, the inequalities $\dN\leq\dDS$ and $\dN\leq\dG$ persist, but the dimensions may differ drastically: the Graph dimension may be infinite while the DS dimension is finite \citep{DBLP:conf/colt/DanielyS14,AsilisHV25}, and the DS dimension may be infinite while the Natarajan dimension equals one; only the DS dimension characterizes learnability in general \citep{DBLP:conf/focs/BrukhimCDMY22}.

\section{Proof of \texorpdfstring{\prettyref{thm:linearvc}}{Theorem \getrefnumber{thm:linearvc}}}

In this section, we provide a complete proof of \prettyref{thm:linearvc}, which we restate below.

\realizablethm*

\subsection{Leave-One-Out Analysis of Bagging}

We first show that, on most test examples, every step along the target chain-of-thought is predicted correctly by most CoT-consistent predictors: for each $t\in[T]$, the probability over the training sample $S\sim P^n$ that $\widehat f_S$ errs on the prefix $x\circ y_{1:t-1}$ is small. This guarantee is per step, and the predictors that err may differ across steps; it is the majority vote that later combines these per-step guarantees into a correct chain-of-thought.

\begin{lemma}
    \label{lem:key}
    For any autoregressive class $\Fc$ with Graph dimension $\dG$, any $\CONS$ oracle, and any distribution $P$ over $\Sigma^\star \times \Sigma^T$ such that $P\brk{f_\star^{(1:T)}(x)=y}=1$ for some $f_\star\in \Fc$,
    \[
        \Ex{(x,y)\sim P}\brk{a(x,y)^2} \leq C\cdot\frac{\dG}{n}, \qquad a(x, y):= \max_{1\leq t\leq T}~~\Pr_{S\sim P^n} \brk{\widehat{f}_S(x\circ y_{1:t-1}) \neq y_t},
    \]
    where $C\geq1$ is the universal constant of \prettyref{lem:vc-moment-2}.
\end{lemma}

The proof relies on the following second-moment bound for consistent learning rules over VC classes. Classical first-moment bounds \citep{DBLP:journals/jacm/BlumerEHW89} would also suffice, at the cost of a multiplicative $\log(n/\dG)$ factor.

\begin{lemma}[\citep{DBLP:conf/colt/Aden-AliHLZ24,DBLP:journals/corr/abs-2606-13614}]
    \label{lem:vc-moment-2}
    There is a universal constant $C\geq1$ such that the following holds. Let $P$ be a distribution on $\X\times\Y$, and let $\mathcal E$ be a class of subsets of $\X\times\Y$ with VC dimension $d$. Let $E$ be a deterministic selector such that $E(S)\in\mathcal E$ for every $S\in(\X\times\Y)^n$ and $\Pr_{S\sim P^n}\brk{E(S)\cap S=\varnothing}=1$. Then,
    \[
        \Ex{(x,y)\sim P}\brk{\Pr_{S\sim P^n}\brk{(x,y)\in E(S)}^{\,2}}\leq C\cdot\frac{d}{n}.
    \]
\end{lemma}

\begin{proof}[Proof of \prettyref{lem:key}]
    The quantity $a(x,y)$ is the error probability of $\widehat f_S$ at the worst step along the target chain-of-thought. The idea is to fix this worst step for each example in advance, which reduces $a(x,y)$ to the error probability on a single next-token example, and then to apply \prettyref{lem:vc-moment-2} to the resulting class of error regions.

    We first fix, for each example, the prefix on which this predictor is most likely to err. For $(x,y)\in\Sigma^\star\times\Sigma^T$, let $t(x,y)\in[T]$ be the smallest index attaining the maximum in the definition of $a(x,y)$, and define the map $\phi\colon\Sigma^\star\times\Sigma^T\to\Sigma^\star\times\Sigma$ by
    \[
        \phi(x,y):=\prn{x\circ y_{1:t(x,y)-1},\,y_{t(x,y)}}.
    \]
    In words, $\phi$ maps a prompt and its target chain-of-thought to a single next-token example: the prefix of the chain-of-thought up to step $t(x,y)-1$, labeled by the correct token at step $t(x,y)$. The map $\phi$ depends on $P$, $n$, and the $\CONS$ oracle, but not on the sample $S$. Writing $\phi=(\phi_1,\phi_2)$, we have by construction
    \begin{equation}
        \label{eqn:single-prefix}
        a(x,y)=\Pr_{S\sim P^n}\brk{\widehat f_S(\phi_1(x,y))\neq\phi_2(x,y)}.
    \end{equation}

    We now verify the conditions of \prettyref{lem:vc-moment-2}: we exhibit a class of subsets of $\Sigma^\star\times\Sigma^T$ of VC dimension at most $\dG$, together with a deterministic selector into this class whose selected set is disjoint from the sample. For each $f\in\Fc$, define the error region relative to $\phi$, and the class of all such regions, by
    \[
        E_f:=\set{(x,y)\in\Sigma^\star\times\Sigma^T:f(\phi_1(x,y))\neq\phi_2(x,y)},
        \qquad
        \mathcal E:=\set{E_f:f\in\Fc}.
    \]
    We have $(x,y)\in E_f$ if and only if $\phi(x,y)\in G_f$, where $G_f:=\set{(x',y')\in\Sigma^\star\times\Sigma:f(x')\neq y'}$ is the token-level error region of $f$. Hence, if $\mathcal E$ shatters $k$ points, then $\set{G_f:f\in\Fc}$ shatters their $k$ images under $\phi$, which are necessarily distinct, and therefore $\vc(\mathcal E)\leq\vc\prn{\set{G_f:f\in\Fc}}=\dG$.

    Next, consider the selector $E(S):=E_{\widehat f_S}$, which is deterministic and takes values in $\mathcal E$. By realizability, there exists $f_\star \in \Fc$ satisfying $\Pr_{(x,y)\sim P}[f_\star^{(1:T)}(x)=y]=1$. By \eqref{eqn:erm} it follows that $\widehat f_S$ is CoT-consistent with the sample $S$; that is, $\widehat f_S(x_j\circ y_{j,1:t-1})=y_{j,t}$ for all $(x_j, y_j)\in S$ and $t\in[T]$. In particular, $\widehat f_S(\phi_1(x_j,y_j))=\phi_2(x_j,y_j)$ for every $j$, so that $E(S)\cap S=\varnothing$ with probability 1.

    We are now ready to invoke \prettyref{lem:vc-moment-2}. By \eqref{eqn:single-prefix}, $a(x,y)=\Pr_{S\sim P^n}\brk{(x,y)\in E(S)}$, so \prettyref{lem:vc-moment-2}, applied to $\mathcal E$ and the selector $E$, yields
    \[
        \Ex{(x,y)\sim P}\brk{a(x,y)^2}
        =\Ex{(x,y)\sim P}\brk{\Pr_{S\sim P^n}\brk{(x,y)\in E(S)}^{\,2}}
        \leq C\cdot\frac{\dG}{n}.\qedhere
    \]
\end{proof}

\paragraph{Bagging.} We introduce notation for bagging. For a sequence $S=((x_1,y_1), \dots, (x_n,y_n))$ and a bootstrap size $m\geq1$, let $S' \sim {\rm Unif}(S)^m$ be a bootstrap of $m$ examples drawn uniformly at random (with replacement) from $S$, and let $\widehat{f}_{S'}=\CONS(S')$ be the predictor returned by the $\CONS$ oracle \eqref{eqn:erm} on $S'$. Equivalently, we can express this as drawing a sequence of indices $I=(i_1,\dots, i_m) \in [n]^m$ uniformly at random, and denote the bootstrap sample by $S(I)=((x_{i_1}, y_{i_1}), \dots, (x_{i_m}, y_{i_m}))$.

Define the bagging predictor $\widehat{\B}^m_S$, which maps each input $x\in\Sigma^\star$ to a distribution $\widehat{\B}^m_S(\cdot\mid x)$ over $\Sigma$, whose bagging score $\widehat{\B}^m_S(y\mid x)$ is the fraction of the $n^m$ ordered bootstraps of $S$ of size $m$ on which the returned predictor outputs $y$ on input $x$:

\begin{equation}
    \label{eqn:full-bagging}
    \widehat{\B}^m_S(y\mid x) := \Pr_{S'\sim {\rm Unif}(S)^m} \brk{\widehat{f}_{S'}(x) = y} = \frac{1}{n^m} \sum_{I \in [n]^m} \1\set{\widehat{f}_{S(I)}(x)=y}.
\end{equation}
Since each predictor outputs exactly one token, $\sum_{y\in\Sigma}\widehat{\B}^m_S(y\mid x)=1$, so $\widehat{\B}^m_S(\cdot\mid x)$ is indeed a conditional distribution over next tokens; on input $x$, the majority vote of all bootstrapped predictors outputs the mode ${\arg\max_{y\in\Sigma}\widehat{\B}^m_S(y\mid x)}$ of this distribution. When the bootstrap size equals the size of its source sample, we use the shorthand $\widehat{\B}_S:=\widehat{\B}^{|S|}_S$.

We now proceed to bound the leave-one-out risk of the bagging predictor \eqref{eqn:full-bagging}.

\begin{lemma}[Leave-One-Out Error]
    \label{lem:leave-one-out-margin}
    For any autoregressive class $\Fc$ with Graph dimension $\dG$, any $\CONS$ oracle, and any realizable sequence $S = ((x_1, y_1), \dots, (x_n, y_n))$ with $n\geq 2$,
    \[
        \frac{1}{n} \sum_{i=1}^{n} \max_{t\in[T]}\prn{1-\widehat{\B}_{S_{-i}}(y_{i,t}\mid x_i\circ y_{i, 1:t-1})}^2 \leq 2e^2C\cdot \frac{\dG}{n}.
    \]
\end{lemma}

\begin{remark}[Expected Risk] \label{rem:expectationbound}
    Note that by an exchangeability argument, applying \prettyref{lem:leave-one-out-margin} to a sample of size $n+1$ yields the expected bound
    \[
        \Ex{S\sim P^n}\Ex{(x,y)\sim P}\brk{\max_{t\in[T]}\prn{1-\widehat{\B}_{S}(y_t\mid x\circ y_{1:t-1})}^2} \leq 2e^2C\cdot\frac{\dG}{n+1}.
    \]
    Since the majority vote of all bootstrapped predictors errs on the chain-of-thought of $(x,y)$ only if the bagging distribution puts at most half of its mass on the correct token at some step, that is, only if $\widehat{\B}_S(y_t\mid x\circ y_{1:t-1})\leq 1/2$ for some step $t\in[T]$, its expected CoT risk \eqref{eqn:cot-risk} is at most $8e^2C\dG/(n+1)$.
\end{remark}

\begin{proof}[Proof of \prettyref{lem:leave-one-out-margin}]
    The idea is that a bootstrap of $S$ of size $n-1$ omits the $i$-th example with probability at least $1/e$, and is then simply a bootstrap of $S_{-i}$. The leave-one-out errors are therefore controlled by the errors of bagging on the full sequence $S$, to which \prettyref{lem:key} applies with $P$ being the uniform distribution over $S$. This is the observation underlying the out-of-bag estimate of the error of bagging \citep{Breiman96OOB}.

    Let $m=n-1$ and $P=\Unif(S)$. Fix $i\in[n]$ and $t\in[T]$, and write $x'=x_i\circ y_{i,1:t-1}$ and $y'=y_{i,t}$ for the corresponding prefix and its next token. Draw $I$ uniformly from $[n]^m$. Conditionally on the event $i\notin I$, which has probability $(1-1/n)^{n-1}\geq 1/e$, the index sequence $I$ is uniform over $([n]\setminus\set{i})^m$, so $S(I)$ is a bootstrap of $S_{-i}$ of size $m=|S_{-i}|$. Hence, by the law of total probability and \eqref{eqn:full-bagging},
    \[
        1-\widehat{\B}^m_S(y'\mid x')
        =\Pr_I\brk{\widehat f_{S(I)}(x')\neq y'}
        \geq\Pr_I\brk{i\notin I}\cdot\Pr_I\brk{\widehat f_{S(I)}(x')\neq y'\mid i\notin I}
        \geq\frac{1}{e}\prn{1-\widehat{\B}_{S_{-i}}(y'\mid x')}.
    \]
    In words, CoT-consistent predictors trained on bootstraps of $S_{-i}$ fail to predict $y'$ on $x'$ at most $e$ times as often as those trained on bootstraps of $S$.

    Now observe that, for every $(x,y)\in\Sigma^\star\times\Sigma^T$,
    \[
        \max_{t\in[T]}\prn{1-\widehat{\B}^m_S(y_t\mid x\circ y_{1:t-1})}
        =\max_{t\in[T]}\Pr_{S'\sim P^m}\brk{\widehat f_{S'}(x\circ y_{1:t-1})\neq y_t}
        =a(x,y)
    \]
    is precisely the quantity of \prettyref{lem:key}, for the distribution $P$ and samples of size $m$. Combining this identity with the comparison above, taking the maximum over $t\in[T]$, and squaring, we obtain, for each $i\in[n]$,
    \[
        \max_{t\in[T]}\prn{1-\widehat{\B}_{S_{-i}}(y_{i,t}\mid x_i\circ y_{i,1:t-1})}^2\leq e^2a(x_i,y_i)^2.
    \]
    Averaging over $i\in[n]$, and noting that the average of $a(x_i,y_i)^2$ over the $n$ examples of $S$ is its expectation under $P=\Unif(S)$, we obtain
    \[
        \frac{1}{n}\sum_{i=1}^{n}\max_{t\in[T]}\prn{1-\widehat{\B}_{S_{-i}}(y_{i,t}\mid x_i\circ y_{i,1:t-1})}^2
        \leq e^2\,\Ex{(x,y)\sim P}\brk{a(x,y)^2}
        \leq e^2C\cdot\frac{\dG}{m},
    \]
    where the last inequality follows from \prettyref{lem:key}, which applies since $P = \Unif(S)$ is realizable because $S$ is. The lemma follows since $m=n-1\geq n/2$ for $n\geq 2$.
\end{proof}

\subsection{High-Probability Bound and Sparsification}

Our goal here is to convert the leave-one-out error bound of \prettyref{lem:leave-one-out-margin} into a high-probability bound, and then approximate the resulting suffix-bagging predictor using finitely many calls to the $\CONS$ oracle.

\paragraph{Suffix Averaging and the Suffix Bagging Predictor.} Our first step is to apply the suffix averaging technique of Aden-Ali et al.~\citep{DBLP:conf/focs/Aden-AliCSZ23} to bagging \eqref{eqn:full-bagging}. For a sequence $S=((x_j,y_j))_{j=1}^n$ with $n\geq 4$, let
\[
    J_n=\set{\ceil{n/4},\dots,n-1},
    \qquad S_{\leq k}=((x_j,y_j))_{j=1}^k \quad\text{for } k\in J_n.
\]
Here $S_{\leq k}$ is a prefix of length $k$. We introduce next the {\em suffix-bagging predictor} which averages the bagging predictors $\widehat{\B}_{S_{\leq k}}$ over suffixes $k\in J_n$ and then proceed to bound its risk with high probability. Formally, define, for $x\in\Sigma^\star$ and $y\in\Sigma$,
\begin{equation}
    \label{eqn:suffix-agg}
    \begin{aligned}
        \widehat{g}_S(y\mid x)
        &:=\frac{1}{|J_n|}\sum_{k\in J_n}\widehat{\B}_{S_{\leq k}}(y\mid x)
        =\Ex{k\sim\Unif(J_n)}\brk{\Pr_{S'\sim\Unif(S_{\leq k})^k}\brk{\widehat{f}_{S'}(x)=y}}\\
        &=\frac{1}{|J_n|}\sum_{k\in J_n}\frac{1}{k^k}\sum_{I\in[k]^k}\1\set{\widehat{f}_{S(I)}(x)=y}.
    \end{aligned}
\end{equation}
As an average of the distributions $\widehat{\B}_{S_{\leq k}}(\cdot\mid x)$, $\widehat{g}_S(\cdot\mid x)$ is again a distribution over $\Sigma$, and $\widehat{g}_S(y\mid x)$ is the probability that the $\CONS$ oracle, run on a random bootstrap of a randomly chosen prefix, predicts token $y$ on input $x$. The next lemma shows that, with high probability over the sample, the squared error of the suffix-bagging vote at the worst step along the target chain-of-thought is small on average over test examples.

\begin{lemma}[High-probability bound]
    \label{lem:full-agg}
    For any autoregressive class $\Fc$ with Graph dimension $\dG$, any $\CONS$ oracle, any distribution $P$ over $\Sigma^\star\times\Sigma^T$ such that
    $P\brk{f_\star^{(1:T)}(x)=y}=1$ for some $f_\star\in\Fc$, any $n\geq 4$, and any $\delta\in(0,1)$, with probability at least $1-\delta$ over $S\sim P^n$,
    \[
        \Ex{(x,y)\sim P}\brk{\max_{t\in[T]}\prn{1-\widehat{g}_S(y_t\mid x\circ y_{1:t-1})}^2}
        \leq 19.28\,e^2C\prn{\frac{\dG}{n}+\frac{1}{n}\log\frac{2}{\delta}}.
    \]
\end{lemma}

\begin{proof}
    The proof has two steps. We first convert the leave-one-out guarantee of \prettyref{lem:leave-one-out-margin} into a high-probability bound on the loss of the bagging predictors, averaged over prefixes $S_{\leq k}$ of the sample; we then pass from this average to the suffix-bagging predictor $\widehat g_S$ by Jensen's inequality.

    The bagging predictor $\widehat{\B}_S^m$ is invariant under permutations of $S$, because it averages over all ordered bootstraps, even when the $\CONS$ oracle depends on the order of its input. Moreover, \prettyref{lem:leave-one-out-margin} bounds the average leave-one-out loss $\frac{1}{n}\sum_{i=1}^n\max_{t\in[T]}(1-\widehat{\B}_{S_{-i}}(y_{i,t}\mid x_i\circ y_{i,1:t-1}))^2$ by $2e^2C\dG/n$, uniformly over all realizable sequences, and this loss takes values in $[0,1]$. We may therefore invoke the suffix-averaging bound of Aden-Ali et al.~\citep[][Theorem 2.1]{DBLP:conf/focs/Aden-AliCSZ23} for this loss:\footnote{For $n$ divisible by $4$, \citep[][Theorem 2.1]{DBLP:conf/focs/Aden-AliCSZ23} bounds the average risk over $k\in\set{n/4,\dots,n-1}$ by $4.82\,(M+\log(2/\delta))/n$, where $M$ bounds the sum of the leave-one-out losses; here $M=2e^2C\dG$. For general $n\geq4$, apply it to the first $n'=4\lfloor n/4\rfloor\geq n/2$ examples. Compared with its index set, $J_n$ omits at most one index and adds at most three, each with loss at most $1$; since $|J_n|\geq\max\set{n/2,3n'/4}$, this changes the bound by at most a factor of two and an additive $6/n$. Since $6\leq 9\log(2/\delta)$ and $C\geq1$, the resulting bound $(19.28\,e^2C\dG+9.64\log(2/\delta)+6)/n$ is at most the right-hand side of \eqref{eqn:suffix-average-risk}.} with probability at least $1-\delta$ over $S\sim P^n$,
    \begin{equation}
        \label{eqn:suffix-average-risk}
        \frac{1}{|J_n|}\sum_{k\in J_n}\Ex{(x,y)\sim P}\brk{\max_{t\in[T]}\prn{1-\widehat{\B}_{S_{\leq k}}(y_t\mid x\circ y_{1:t-1})}^2}\leq 19.28\,e^2C\cdot\frac{\dG+\log(2/\delta)}{n}.
    \end{equation}

    It remains to pass from the average loss of the bagging predictors $\widehat{\B}_{S_{\leq k}}$ to the loss of their average $\widehat g_S$. Fix $(x,y)\in\Sigma^\star\times\Sigma^T$, and write $x_t=x\circ y_{1:t-1}$ for $t\in[T]$. By \eqref{eqn:suffix-agg}, $\widehat g_S(y_t\mid x_t)$ is the average of $\widehat{\B}_{S_{\leq k}}(y_t\mid x_t)$ over $k\in J_n$, so Jensen's inequality gives, for every $t\in[T]$,
    \[
        \prn{1-\widehat g_S(y_t\mid x_t)}^2\leq\frac{1}{|J_n|}\sum_{k\in J_n}\prn{1-\widehat{\B}_{S_{\leq k}}(y_t\mid x_t)}^2\leq\frac{1}{|J_n|}\sum_{k\in J_n}\max_{s\in[T]}\prn{1-\widehat{\B}_{S_{\leq k}}(y_s\mid x_s)}^2.
    \]
    The right-hand side does not depend on $t$. Taking the maximum over $t\in[T]$ on the left-hand side, then expectations over $(x,y)\sim P$, and applying \eqref{eqn:suffix-average-risk} proves the claim.
\end{proof}

The suffix-bagging predictor \eqref{eqn:suffix-agg} averages over all ordered bootstraps and is therefore intractable. We approximate its token votes by sampling finitely many predictors, as in \prettyref{alg:bagging}. The following sparsification bound controls the CoT risk \eqref{eqn:cot-risk} directly, in the spirit of \citep[][Lemma~3]{DBLP:conf/colt/Larsen23}. The idea is simple: the sampled majority errs at a target prefix only if at least half of the sampled predictors err there, an event whose probability decays exponentially in $N$ and is at most proportional to the squared error of the vote, controlled by \prettyref{lem:full-agg}. Since this quantity appears both in the bound on the expected error and in the threshold of Markov's inequality, it cancels from the failure probability, and the number of oracle calls does not depend on $n$.

\begin{lemma}[Sparsification via binomial tails]
    \label{lem:sparsification-risk}
    Fix a realizable sample $S$ of size $n\geq4$ and a distribution $P$ over $\Sigma^\star\times\Sigma^T$. Let $\widehat h=\MAJ(\widehat{f}_{S'_1},\dots,\widehat{f}_{S'_N})$ be the majority vote, with fixed tie-breaking, of the predictors returned by the $\CONS$ oracle on $N$ independent bootstrap samples drawn as in steps 3--4 of \prettyref{alg:bagging}. For any $\delta\in(0,1)$, if $N\geq\max\set{4,\log_2(T/\delta)}$, then, with probability at least $1-\delta$ over the bootstrap samples,
    \begin{equation}
        \label{eqn:risk-sparsification}
        \Pr_{(x,y)\sim P}\brk{\exists t\in[T]: \widehat h^{(t)}(x)\neq y_t}\leq 256\,\Ex{(x,y)\sim P}\brk{\max_{t\in[T]}\prn{1-\widehat{g}_S(y_t\mid x\circ y_{1:t-1})}^2}.
    \end{equation}
\end{lemma}

\begin{proof}
    Fix $S$. Throughout the proof, probabilities and expectations without a subscript are over the bootstrap samples $S'_1,\dots,S'_N$. Write
    \[
        r(x,y):=\max_{t\in[T]}\prn{1-\widehat g_S(y_t\mid x\circ y_{1:t-1})}^2
    \]
    for the squared worst-step error of the suffix-bagging predictor on $(x,y)\in\Sigma^\star\times\Sigma^T$, so that the right-hand side of \eqref{eqn:risk-sparsification} is $256\,\E_{(x,y)\sim P}[r(x,y)]$. Let $E$ be the set of examples with $r(x,y)\geq 1/256$, that is, with $\widehat g_S(y_t\mid x\circ y_{1:t-1})\leq 15/16$ for some $t\in[T]$. Then, for every realization of the bootstrap samples,
    \[
        \Pr_{(x,y)\sim P}\brk{\exists t\in[T]: \widehat h^{(t)}(x)\neq y_t}\leq P(E)+Z,
        \qquad
        Z:=\Pr_{(x,y)\sim P}\brk{(x,y)\notin E,\ \exists t\in[T]: \widehat h^{(t)}(x)\neq y_t}.
    \]
    Since $r(x,y)\geq 1/256$ on $E$, $P(E)\leq 256\,\E_{(x,y)\sim P}[\1\set{(x,y)\in E}\,r(x,y)]$. It suffices to show that, with probability at least $1-\delta$, $Z\leq 256\,\E_{(x,y)\sim P}[\1\set{(x,y)\notin E}\,r(x,y)]$; adding the two bounds proves \eqref{eqn:risk-sparsification}.

    By linearity of expectation,
    \[
        \E\brk{Z}=\Ex{(x,y)\sim P}\brk{\1\set{(x,y)\notin E}\,\Pr\brk{\exists t\in[T]: \widehat h^{(t)}(x)\neq y_t}}.
    \]
    Now fix $(x,y)\notin E$. By the definition of $E$, we have $r(x,y)<1/256$, that is, $\max_{t\in[T]}(1-\widehat g_S(y_t\mid x\circ y_{1:t-1}))<1/16$, so the suffix-bagging score of the correct token exceeds $15/16$ at every step. Since $\widehat h$ follows the target chain-of-thought until its first mistake, it errs on $(x,y)$ only if $\widehat h(x\circ y_{1:t-1})\neq y_t$ for some $t\in[T]$. Fix a step $t$, and let $p=1-\widehat g_S(y_t\mid x\circ y_{1:t-1})<1/16$. We claim that the mistakes of the sampled predictors at this step are independent Bernoulli variables with mean $p$. Indeed, the bootstrap samples $S'_1,\dots,S'_N$ are independent, and each $S'_j$ is drawn exactly as in \eqref{eqn:suffix-agg}, by choosing $k$ uniformly from $J_n$ and then a uniform bootstrap of $S_{\leq k}$. Hence the indicators $\1\set{\widehat f_{S'_j}(x\circ y_{1:t-1})\neq y_t}$, for $j\in[N]$, are independent, and each has mean
    \[
        \Pr\brk{\widehat f_{S'_j}(x\circ y_{1:t-1})\neq y_t}=1-\widehat g_S(y_t\mid x\circ y_{1:t-1})=p.
    \]
    If fewer than half of the sampled predictors err, then $y_t$ receives a strict majority of the votes. Hence $\widehat h(x\circ y_{1:t-1})\neq y_t$ only if some set of $\ceil{N/2}$ predictors all err, an event of probability $p^{\ceil{N/2}}\leq p^{N/2}$ for each fixed set by independence. A union bound over the at most $2^N$ such sets gives, using $4p\leq 1/4$ and $N\geq4$,
    \[
        \Pr\brk{\widehat h(x\circ y_{1:t-1})\neq y_t}\leq 2^Np^{N/2}=(4p)^2\,(4p)^{N/2-2}\leq 16p^2\cdot 4^{2-N/2}=256\,p^2\,2^{-N}.
    \]
    A union bound over the $T$ steps yields $\Pr\brk{\exists t\in[T]: \widehat h^{(t)}(x)\neq y_t}\leq 256\,T2^{-N}r(x,y)$ for every $(x,y)\notin E$. Plugging this into the expression for $\E\brk{Z}$, and $T2^{-N}\leq\delta$ (as $N\geq\log_2(T/\delta)$), we obtain
    \[
        \E\brk{Z}\leq 256\,\delta\,\Ex{(x,y)\sim P}\brk{\1\set{(x,y)\notin E}\,r(x,y)}.
    \]
    Markov's inequality now implies that, with probability at least $1-\delta$,
    \[
        Z\leq 256\,\Ex{(x,y)\sim P}\brk{\1\set{(x,y)\notin E}\,r(x,y)},
    \]
    where, if this expectation is zero, then $Z=0$ almost surely. This completes the proof.
\end{proof}

Alternatively, the sparsification argument of Moran and Yehudayoff~\citep{DBLP:journals/jacm/MoranY16} approximates all suffix-bagging scores simultaneously, with a number of samples independent of $T$ but depending on the dual VC dimension of $\Fc$. The following high-probability form follows by uniform convergence for the dual class of $\Fc$.

\begin{lemma}[Sparsification via dual VC dimension]
    \label{lem:sparsification}
    Let $\Fc$ be a class of functions from $\Sigma^\star$ to $\Sigma$. For each $(x,y)\in\Sigma^\star\times\Sigma$, define the evaluation function $g_{(x,y)}\colon\Fc\to\set{0,1}$ by $g_{(x,y)}(f):=\1\set{f(x)=y}$. The dual class of $\Fc$ is $\mathcal G:=\set{g_{(x,y)}:(x,y)\in\Sigma^\star\times\Sigma}$, and its VC dimension $d^\star:=\vc(\mathcal G)$ is the dual VC dimension of $\Fc$. Let $Q$ be any distribution over $\Fc$. For any $\eps,\delta\in(0,1)$, with probability at least $1-\delta$, $N=O\prn{(d^\star+\log(1/\delta))/\eps^2}$ independent predictors $f_1,\dots,f_N\sim Q$ satisfy
    \[
        \sup_{(x,y)\in\Sigma^\star\times\Sigma}\ \left|\frac{1}{N}\sum_{j=1}^N g_{(x,y)}(f_j)-\Ex{f\sim Q}\brk{g_{(x,y)}(f)}\right|\leq\eps.
    \]
\end{lemma}

\subsection{Putting the Pieces Together}

\begin{proof}[Proof of \prettyref{thm:linearvc}]
    Let $S\sim P^n$, and let $\widehat h=\MAJ(\widehat{f}_{S'_1},\dots,\widehat{f}_{S'_N})$ be the output of Algorithm~\ref{alg:bagging}. By \prettyref{lem:full-agg}, with probability at least $1-\delta/2$ over $S$,
    \begin{equation}
        \label{eqn:bound1}
        \Ex{(x,y)\sim P}\brk{\max_{t\in[T]}\prn{1-\widehat{g}_S(y_t\mid x\circ y_{1:t-1})}^2}
        \leq 19.28\,e^2C\prn{\frac{\dG}{n}+\frac{1}{n}\log\frac{4}{\delta}}.
    \end{equation}
    For every fixed $S$, our choice $N=\max\set{4,\ceil{\log_2(2T/\delta)}}$ allows us to apply \prettyref{lem:sparsification-risk} with failure probability $\delta/2$. Thus, with conditional probability at least $1-\delta/2$, the CoT risk \eqref{eqn:cot-risk} of $\widehat h$ is at most $256$ times the left-hand side of \eqref{eqn:bound1}. Combining the two guarantees by a union bound, with probability at least $1-\delta$ over the training and bootstrap samples,
    \[
        \Pr_{(x,y)\sim P}\brk{\exists t\in[T]: \widehat h^{(t)}(x)\neq y_t}
        \leq 256\cdot19.28\,e^2C\prn{\frac{\dG}{n}+\frac{1}{n}\log\frac{4}{\delta}}.
    \]
    For $\dG\geq1$, we have $\dG+\log(4/\delta)\leq(1+\log4)(\dG+\log(1/\delta))$, giving the stated bound with constant $256\cdot19.28\,(1+\log 4)\,e^2C$. When $\dG=0$, all functions in $\Fc$ agree on every next-token, and the CoT risk \eqref{eqn:cot-risk} is zero. The algorithm makes $N=O\prn{\log(T/\delta)}$ calls to the $\CONS$ oracle, each on a bootstrap of fewer than $n$ labeled examples.
\end{proof}

\section{Distilling the Ensemble into a Single Generator}
\label{sec:distillation}

\prettyref{alg:bagging} achieves the optimal, $T$-independent sample complexity of \prettyref{thm:linearvc} by aggregating an ensemble of $N=\max\set{4,\ceil{\log_2(2T/\delta)}}$ generators in $\Fc$. Although the ensemble is efficient to train, it must store all $N$ generators and evaluate each of them at every step of generation, multiplying the computational cost per token and the memory by $N$. This becomes costly when the generators are frontier-scale language models. Current frontier open-weight models have hundreds of billions to trillions of parameters \citep{GLMTeam26,DeepSeekAI26,KimiTeam26K3}, so a single copy requires terabytes of memory to store and must be sharded across many GPUs to run. Moreover, the ensemble size $N$ grows with the CoT length $T$, which is large and growing rapidly in practice. Context windows now reach millions of tokens \citep{BurnhamAdamczewski2025ContextWindows}, and the length of reasoning traces is estimated to increase by about $5 \times$ per year \citep{Emberson2025OutputLength}.

We show that the ensemble can be distilled into a single generator in the class $\Fc$ after training, using only unlabeled prompts. The procedure draws $\nU$ fresh \emph{unlabeled} prompts, labels them with the chain-of-thought traces generated by the majority-vote ensemble, and fits a single generator in $\Fc$ to those traces via empirical risk minimization. This is an instance of sequence-level distillation \citep{KimR16}. 
Distillation is a standard technique for compressing a large model or an ensemble into a smaller one \citep{BuciluaCaruanaNiculescuMizil06,HintonVD15}, and it is now widely used in language model training, where smaller models are trained on chains-of-thought generated by a large reasoning model \citep[e.g.,][]{QwenTeam25,GuoYZSWZXZMBZYW25}. With $\nU \geq n \log(n T)$ unlabeled prompts (which are much cheaper to obtain than chain-of-thought demonstrations), the distilled generator retains the $O((\dG + \log(1/\delta)) / n)$ CoT risk of \prettyref{thm:linearvc}.

\paragraph{Setting.}
Let $\widehat h = \MAJ(\widehat f_{S'_1},\dots,\widehat f_{S'_N})$ be the majority-vote ensemble returned by \prettyref{alg:bagging}. Let $P_X$ denote the prompt marginal of the distribution $P$ over $\Sigma^\star \times \Sigma^T$ realizable by $f_\star \in \Fc$. Distillation uses $\nU$ fresh prompts $x_1', \ldots, x_{\nU}' \sim P_X$, drawn independently of $\widehat h$, whose target chains-of-thought are not observed. Instead, each prompt is labeled by $\widehat h$, which acts as the \emph{teacher}, labeling each prompt with its own chain-of-thought $y_i' \coloneqq \widehat h^{(1:T)}(x_i')$. This gives the teacher-labeled sample $\bar S = ((x_i', y_i'))_{i=1}^{\nU}$, drawn i.i.d.\ from the distribution $P_{\widehat h}$ of $(x,\widehat h^{(1:T)}(x))$ for $x\sim P_X$.

Since the majority-vote ensemble $\widehat h$ need not be in $\Fc$, the teacher-labeled sample $\bar S$ may not be realizable by $\Fc$. Thus, instead of a consistency oracle $\CONS$, we assume an $\ERMCoT$ oracle, which returns a generator in $\Fc$ that minimizes the empirical CoT risk. The distillation process produces a \emph{student} $\bar f$ by running $\ERMCoT$ on $\bar S$:
\begin{equation}
    \label{eqn:erm-cot}
    \bar f = \ERMCoT(\bar S) \in \argmin_{f \in \Fc} \frac{1}{\nU} \sum_{i=1}^{\nU} \1\set{\exists t\in[T]:\ f^{(t)}(x_i')\neq y'_{i,t}}.
\end{equation}

The following proposition shows that the student $\bar f$, a single generator in $\Fc$, retains the rate of \prettyref{thm:linearvc}.

\begin{proposition}[Distillation to one generator]
    \label{prop:distill}
    Let $P$ be realizable by $f_\star\in\Fc$, and let $\widehat h$ be any generator with $\Pr_{(x,y)\sim P}[\exists t\in[T]:\widehat h^{(t)}(x)\neq y_t]\le\hat\eps$. Let $\bar f$ be the student \eqref{eqn:erm-cot} trained on $\nU$ fresh prompts $x_1',\dots,x_{\nU}' \sim P_X$, drawn independently of $\widehat h$ and labeled by $\widehat h$. For every $\delta\in(0,1)$, with probability at least $1-\delta$ over the prompts,
    \begin{equation}\label{eqn:prop:distill-1}
        \Pr_{(x,y)\sim P}\brk{\exists t\in[T]:\ \bar f^{(t)}(x)\neq y_t} \leq 5 \hat\eps + \frac{9}{\nU} \prn{\dG \log(2 \nU T + 1)+ \log \frac{8}{\delta}}.
    \end{equation}
    Consequently, if $\widehat h = \MAJ(\widehat f_{S'_1},\dots,\widehat f_{S'_N})$ is the output of \prettyref{alg:bagging} run with confidence $\delta$ on $S\sim P^n$ with $n\geq4$, and $\bar f$ is trained on $\nU \geq n \log(nT)$ fresh prompts, then with probability at least $1 - 2\delta$ over $S$, the bootstrap samples, and the unlabeled prompts, we have
    \begin{equation}\label{eqn:prop:distill-2}
        \Pr_{(x,y)\sim P}\brk{\exists t\in[T]:\ \bar f^{(t)}(x)\neq y_t}=O\prn{\frac{\dG+\log(1/\delta)}{n}}.
    \end{equation}
\end{proposition}

The distillation process trades unlabeled data and a one-time computational cost after training for an $N$-fold reduction in inference compute and memory. \prettyref{prop:distill} retains the labeled sample complexity of \prettyref{thm:linearvc}, which is optimal and independent of $T$. In \prettyref{thm:linearvc}, the dependence on $T$ appears only in the ensemble size $N$, paid at every inference. Distillation moves the dependence on $T$ to the number of unlabeled prompts $\nU \geq n \log(nT)$, paid once. Although the teacher-labeled distribution $P_{\widehat h}$ need not be realizable by $\Fc$, it is nearly realizable, since $f_\star$ and $\widehat h$ disagree only where $\widehat h$ errs, which by \prettyref{thm:linearvc} happens with probability at most $O((\dG + \log(1/\delta))/n)$. The relative deviation inequalities of \prettyref{lem:reldev} exploit this to give the fast $1/\nU$ rate in \eqref{eqn:prop:distill-1}, where a worst-case agnostic analysis would give only $1/\sqrt{\nU}$. A similar compression-via-distillation procedure appears in \citep{UrnerShalevShwartzBenDavid11}, where unlabeled inputs, labeled by an accurate but expensive predictor, are used to train a predictor from a cheaper class. Their analysis is agnostic because their student class may have approximation error, so their bound on the number of unlabeled samples scales as $1/\eps^2$ rather than $1/\eps$. 

\subsection{Proof of \texorpdfstring{\prettyref{prop:distill}}{Proposition \getrefnumber{prop:distill}}}
\label{sec:distillation-proof}

For a generator $f$ and a distribution $D$ over $\Sigma^\star\times\Sigma^T$, write $\mathrm{err}_D(f):=\Pr_{(x,y)\sim D}\brk{\exists t\in[T]:\ f^{(t)}(x)\neq y_t}$ for the CoT risk \eqref{eqn:cot-risk} of $f$ under $D$, and $\widehat{\mathrm{err}}(f):=\frac{1}{\nU}\sum_{i=1}^{\nU}\1\set{\exists t\in[T]:\ f^{(t)}(x_i')\neq y'_{i,t}}$ for its empirical CoT risk on the teacher-labeled sample $\bar S$, which the student minimizes in \eqref{eqn:erm-cot}. The proof relies on the following relative deviation bound for classes of binary functions, stated in terms of the growth function.

\begin{lemma}[{\citep[][Theorem~5.1 and Section~5.1.2]{BoucheronBousquetLugosi05}}]
    \label{lem:reldev}
    Let $\mathcal L$ be a class of $\set{0,1}$-valued functions on a set $\mathcal Z$, with growth function $\Growth{\mathcal L}(k):=\max_{z_1,\dots,z_k\in\mathcal Z}\left|\set{(\ell(z_1),\dots,\ell(z_k)):\ell\in\mathcal L}\right|$. Let $D$ be a distribution over $\mathcal Z$, and let $z_1,\dots,z_m\sim D^m$. For every $\delta\in(0,1)$, with probability at least $1-\delta$, every $\ell\in\mathcal L$ satisfies
    \[
        \Ex{z\sim D}\brk{\ell(z)}\leq\frac{2}{m}\sum_{i=1}^m \ell(z_i)+5\cdot\frac{\log\Growth{\mathcal L}(2m)+\log(4/\delta)}{m}.
    \]
\end{lemma}

\begin{proof}
    This is a standard consequence of relative deviation inequalities, and we include its short derivation for completeness. Write $\gamma:=(\log\Growth{\mathcal L}(2m)+\log(4/\delta))/m$, $A:=\E_{z\sim D}[\ell(z)]$ and $B:=\frac1m\sum_{i=1}^m \ell(z_i)$. By \citep[][Theorem~5.1]{BoucheronBousquetLugosi05}, with probability at least $1-\delta$, every $\ell\in\mathcal L$ satisfies $A\leq B+2\sqrt{A\,\gamma}$. For $A,B,C\geq0$, the inequality $A\leq B+C\sqrt A$ implies $A\leq B+C^2+C\sqrt B$, so $A\leq B+4\gamma+2\sqrt{B\,\gamma}\leq 2B+5\gamma$, where the last step uses $2\sqrt{B\,\gamma}\leq B+\gamma$.
\end{proof}

We apply \prettyref{lem:reldev} to the CoT loss class $\mathcal L:=\set{(x,y)\mapsto\1\set{\exists t\in[T]:\ f^{(t)}(x)\neq y_t}:f\in\Fc}$ over $\Sigma^\star\times\Sigma^T$, whose growth function is controlled by the Graph dimension of $\Fc$.

\begin{lemma}
    \label{lem:growth-cot}
    For any autoregressive class $\Fc$ with Graph dimension $\dG$ and any $k\geq1$, $\Growth{\mathcal L}(k)\leq(kT+1)^{\dG}$.
\end{lemma}

\begin{proof}
    Since a generator follows the target chain-of-thought until its first mistake, $f^{(t)}(x)\neq y_t$ for some $t\in[T]$ if and only if $f(x\circ y_{1:t-1})\neq y_t$ for some $t\in[T]$. Hence, as in \citep[][Theorem~3.4]{pmlr-v291-joshi25a}, the pattern that $\mathcal L$ realizes on $(x_1,y_1),\dots,(x_k,y_k)$ is determined by the pattern of token-level errors of $f$ on the $kT$ next-token examples $(x_i\circ y_{i,1:t-1},y_{i,t})$, for $i\in[k]$ and $t\in[T]$. Recall from \prettyref{sec:prelim} that $\dG$ is the VC dimension of the token-level loss class $\set{(x',y')\mapsto\1\set{f(x')\neq y'}:f\in\Fc}$, so by the Sauer--Shelah lemma \citep{sauer1972density}, it realizes at most $\sum_{i=0}^{\dG}\binom{kT}{i}\leq(kT+1)^{\dG}$ patterns on these $kT$ examples.
\end{proof}

\begin{proof}[Proof of \prettyref{prop:distill}]
    The student is trained on chains-of-thought generated by the teacher $\widehat h$ rather than by $f_\star$, so its training distribution $P_{\widehat h}$ need not be realizable by $\Fc$. The idea is that $P_{\widehat h}$ differs from $P$ only on prompts where $\widehat h$ errs, so $f_\star$ is nearly consistent with $\bar S$, with empirical CoT risk $O(\hat\eps)$. The student $\bar f$ minimizes the empirical CoT risk, so it inherits this small empirical risk from $f_\star$, and \prettyref{lem:reldev} converts it into a small risk, at the fast rate $1/\nU$.

    We first relate the two distributions. Since $y=f_\star^{(1:T)}(x)$ under $P$ and $y=\widehat h^{(1:T)}(x)$ under $P_{\widehat h}$, every generator $f$ satisfies $\mathrm{err}_P(f)=\Pr_{x\sim P_X}[f^{(1:T)}(x)\neq f_\star^{(1:T)}(x)]$ and $\mathrm{err}_{P_{\widehat h}}(f)=\Pr_{x\sim P_X}[f^{(1:T)}(x)\neq\widehat h^{(1:T)}(x)]$. The two events differ only on prompts where $\widehat h^{(1:T)}(x)\neq f_\star^{(1:T)}(x)$, so
    \begin{equation}
        \label{eqn:distill-shift}
        \left|\mathrm{err}_P(f)-\mathrm{err}_{P_{\widehat h}}(f)\right|\leq\mathrm{err}_P(\widehat h)\leq\hat\eps,
        \qquad\text{and in particular}\qquad
        \mathrm{err}_{P_{\widehat h}}(f_\star)=\mathrm{err}_P(\widehat h)\leq\hat\eps.
    \end{equation}

    Next, we bound the empirical CoT risk of $f_\star$. Since the prompts are drawn independently of $\widehat h$, conditionally on $\widehat h$ the indicators $\1\set{\exists t\in[T]:\ f_\star^{(t)}(x_i')\neq y'_{i,t}}$, for $i\in[\nU]$, are independent Bernoulli variables with mean $p:=\mathrm{err}_{P_{\widehat h}}(f_\star)\leq\hat\eps$. By Bernstein's inequality \citep[][Section~5.1.2]{BoucheronBousquetLugosi05}, with probability at least $1-\delta/2$,
    \[
        \widehat{\mathrm{err}}(f_\star)\leq p+\sqrt{\frac{2p\log(2/\delta)}{\nU}}+\frac{2\log(2/\delta)}{3\nU}\leq 2\hat\eps+\frac{2\log(2/\delta)}{\nU},
    \]
    where the last step uses $\sqrt{2pL}\leq p+L/2$ for $L=\log(2/\delta)/\nU$.

    We now apply \prettyref{lem:reldev} to $\mathcal L$, with $D=P_{\widehat h}$, the sample $\bar S$, and confidence $\delta/2$. By \prettyref{lem:growth-cot}, with probability at least $1-\delta/2$, every $f\in\Fc$ satisfies
    \[
        \mathrm{err}_{P_{\widehat h}}(f)\leq 2\,\widehat{\mathrm{err}}(f)+5\gamma,
        \qquad
        \gamma:=\frac{\dG\log(2\nU T+1)+\log(8/\delta)}{\nU}.
    \]
    By a union bound, both bounds hold with probability at least $1-\delta$. On this event, applying \eqref{eqn:distill-shift} to $\bar f$, then the last display, then $\widehat{\mathrm{err}}(\bar f)\leq\widehat{\mathrm{err}}(f_\star)$, which holds since $\bar f$ minimizes the empirical CoT risk over $\Fc$ and $f_\star\in\Fc$, and finally the bound on $\widehat{\mathrm{err}}(f_\star)$, we obtain
    \begin{align*}
        \mathrm{err}_P(\bar f)
        &\leq\mathrm{err}_{P_{\widehat h}}(\bar f)+\hat\eps
        \leq 2\,\widehat{\mathrm{err}}(f_\star)+5\gamma+\hat\eps\\
        &\leq 5\hat\eps+\frac{4\log(2/\delta)}{\nU}+5\gamma
        \leq 5\hat\eps+\frac{9}{\nU}\prn{\dG\log(2\nU T+1)+\log\frac8\delta},
    \end{align*}
    which implies \eqref{eqn:prop:distill-1}.

    For the second statement, by \prettyref{thm:linearvc}, with probability at least $1-\delta$ over $S$ and the bootstrap samples, $\hat\eps=O\prn{(\dG+\log(1/\delta))/n}$. Since the prompts are independent of $\widehat h$, the first statement applies conditionally on $\widehat h$, and a union bound gives both events with probability at least $1-2\delta$. Finally, since $x\mapsto\log(3xT)/x$ is decreasing for $x\geq1$, the assumption $\nU\geq n\log(nT)$ and $\log(nT)\geq\log 4$ give
    \[
        \frac{\log(2\nU T+1)}{\nU}\leq\frac{\log(3\nU T)}{\nU}\leq\frac{\log3+\log(nT)+\log\log(nT)}{n\log(nT)}\leq\frac3n,
    \]
    and $\log(8/\delta)/\nU\leq\log(8/\delta)/n$, so the second term in \eqref{eqn:prop:distill-1} is $O\prn{(\dG+\log(1/\delta))/n}$.
\end{proof}

\section{Additional Related Work}
\label{sec:related}

\paragraph{Learning Theory of Autoregressive Chain-of-Thought.} An early framework for learning from chain-of-thought data was proposed by Malach~\citep{Malach24}, in which each time step is learned separately, with guarantees growing linearly in the CoT length $T$. Later, Joshi et al.~\citep{pmlr-v291-joshi25a} introduced the time-invariant framework we adopt, and showed that the dependence on the length is mild: logarithmic in $T$ under chain-of-thought supervision, and in general linear under end-to-end supervision, where only the final answer is observed. In related work, Altabaa, Montasser, and Lafferty~\citep{NEURIPS2025_235a9abb} introduced the CoT information, which measures the discriminative power gained by observing the chain-of-thought. More recently, Hanneke, Mehalel, and Moran~\citep{HannekeMehalelMoran2026} removed the dependence on $T$ under chain-of-thought supervision, at the cost of a factor of the dual VC dimension, and showed that under end-to-end supervision essentially any growth rate between constant and linear in $T$ can arise. In the online setting, Doron-Arad, Mehalel, and Mossel~\citep{DoronAradMehalelMossel2026} obtained bounds independent of $T$, in terms of the Littlestone dimension. Finally, Li~\citep{Li2026Autoregressive} determined the optimal sample complexity, which scales with the DS dimension. In addition, Yang, Srebro, and Li~\citep{YangSL26} gave tight bounds on the VC dimension of transformers. 

\paragraph{Stochastic Autoregressive Chain-of-Thought.} A parallel line of work studies stochastic generators learned by next-token prediction with the logarithmic loss, and asks how the sequence-level error scales with length. In imitation learning, Foster, Block, and Misra~\citep{FosterBM24} showed that behavior cloning with the logarithmic loss can achieve horizon-independent sample complexity. Under misspecification, the error of next-token prediction relative to the best model must grow with the sequence length for a broad family of objectives \citep{RohatgiBHKF25}, although measuring error in the joint KL divergence avoids this growth in the approximation factor \citep{XuYZ26}. Doron-Arad, Mehalel, and Mossel~\citep{DoronAradMehalelMossel2026Stochastic} extended the framework of Joshi et al.~\citep{pmlr-v291-joshi25a} to binary stochastic generators under the squared loss. In contrast, we study deterministic generators in the realizable setting, where the target CoT must be reproduced exactly.

\paragraph{Verifiers, Rewards, and Curricula.} Beyond learning from demonstrations, Balcan et al.~\citep{BalcanBLS25} introduced a PAC framework for learning verifiers of chains-of-thought, later extended to an online setting by Balcan et al.~\citep{BalcanBFLS26}. In reinforcement learning, Jia, Rakhlin, and Xie~\citep{JiaRX25} showed that outcome supervision is not statistically harder than process supervision, up to polynomial factors in the horizon. Recently, Rajaraman et al.~\citep{RajaramanHDSFK26} showed that an autocurriculum, which focuses supervision on prompts the learner gets wrong, can require exponentially fewer demonstrations.

\paragraph{Expressivity of Chain-of-Thought.} Intermediate generation increases the expressive power of transformers: they become Turing-complete \citep{PerezBM21}, can solve arithmetic and dynamic-programming tasks that bounded-depth transformers cannot solve directly \citep{FengZGYHW23}, and gain the serial computation that constant-depth transformers otherwise lack, in proportion to the number of intermediate steps \citep{MerrillS24,LiLZM24}. Even a single autoregressive neuron can simulate any feedforward circuit with the same activation, while remaining efficiently learnable from chain-of-thought supervision \citep{MagenJV26}. These results concern what can be represented, and complement the statistical question of how many demonstrations are needed to learn it.

\paragraph{Gradient-Based Learning and Length Generalization.} Intermediate supervision can also make problems tractable for gradient-based training: suitable scratchpads overcome a globality barrier for transformers \citep{AbbeBLSS24}, and transformers trained with intermediate steps provably learn parity \citep{KimS25} and state-tracking tasks \citep{HuangWSCC25} efficiently. Conversely, training by next-token prediction with teacher forcing can fail to learn an accurate predictor on simple planning tasks \citep{BachmannN24}. A related question is length generalization, the ability to extrapolate to inputs or chains-of-thought longer than those seen in training \citep{HuangYBSKZNH25,GolowichJBKM25,HuangWSCC25}; in our setting, training and test chains-of-thought have the same length $T$.

\section*{AI Disclosure}
At a technical level, the main idea of this paper is to study the problem of autoregressive learning by adapting and extending the techniques of Montasser~\citep{Montasser26} devised for adversarially robust learning. This was conceived by the authors without the use of AI tools. Along the way, ChatGPT helped us remove a $\log n$ factor from the number of oracle calls (and consequently number of generators $N$) in an earlier version of \prettyref{lem:sparsification-risk}, and Claude helped us polish the writing and presentation.

\microtypesetup{protrusion=false}
\bibliographystyle{surnameinitials}
\bibliography{refs}

\end{document}